# An Online Learning Algorithm for Neuromorphic Hardware Implementation

Chetan Singh Thakur, Runchun Wang, Saeed Afshar, Gregory Cohen, Tara Julia Hamilton, Jonathan Tapson and André van Schaik
Email: chetansingh84@gmail.com

*Abstract*— In this paper, we propose a Sign-based Online Update Learning (SOUL) algorithm, which may be used in any artificial neural network that learns weights by minimising a convex cost function. The SOUL algorithm is a simple weight update rule that employs the sign of the hidden layer activation and the sign of the output error, which is the difference between the observed output and the predicted output. This algorithm is easily implementable in hardware using simply a counter and an XOR gate. Here, we present results of using SOUL to train an analogue Integrated Circuit implementation of the Extreme Learning Machine (ELM) for various regression tasks. We also present results for a Field Programmable Gate Array (FPGA) implementation of a digital ELM system trained using the SOUL algorithm on the MNIST handwritten digit database, and demonstrate its ability to perform digit recognition tasks in real time. The accuracy of the SOUL algorithm in the digit recognition task is lower than state-of-the-art machine learning algorithms, however, the SOUL learning rule is extremely simple and area efficient for hardware implementation. We envisage that it will find applications in low power hardware accelerators for big data machine learning applications.

## I. Introduction

Most learning algorithms developed for neural networks are designed, tested and benchmarked in software on sequential processors and optimised either for performance or for biological plausibility and not for their ease of implementation in custom hardware [1]–[6]. Machine learning for analysing big data is one of the most important workloads in both data centres and mobile platforms. Even state-of-the-art CPUs/GPUs are not sufficiently fast to perform real-time processing in big data applications and their power consumption is a major concern [7]. This gap in the design space serves as a motivation for an integrated algorithm-hardware design paradigm that takes full advantage of the hardware characteristics to implement more efficient and better performing hardware-based neural network systems. In this context, we propose an online learning algorithm called the Sign-based Online Update Learning (SOUL) algorithm, which is optimised for hardware implementation. We have derived the SOUL algorithm from the OPIUM algorithm proposed in our previous work [8]. Incorporating online learning algorithms for real-time applications imposes speed and memory constraints that can only be met by implementing simple learning architectures on silicon. The algorithm is very simple and can be easily implemented using standard digital logic cells, requiring only counters and XOR gates.

In this paper, we implement the SOUL algorithm to perform generic function regression and classification tasks in ELM-based networks. We tested the performance of the algorithm for regression tasks in an analogue system referred as a TAB (Trainable Analogue Block) [9][10]. For classification tasks, we employed a previously developed digital system [11] for the MNIST digit recognition task [12].

This paper is organised as follows: section II describes the ELM framework, we present the OPIUM learning algorithm in section III, the derivation of the SOUL algorithm from OPIUM in section IV, and hardware implementation of the SOUL algorithm in section V. MNIST digit recognition results are shown in section VI, simulation learning results of the TAB framework implemented in Python and transistor circuits are presented in section VII, and conclusions in section VIII.

## II. ELM Framework

An ELM framework [13] consists of a feed-forward structure of three layers of neurons – input, hidden, and output (Fig. 1). An input is projected to a layer of nonlinear hidden neurons of a much higher dimensionality via random connection weights. Additionally, we have introduced a fixed and distinct systematic offset (Fig. 1), $o_i$, for each hidden layer neuron of the ELM framework, which ensures that all the neuronal tuning curves are distinct and independent. For the analogue TAB system used in this work, the hidden neurons employ a hyperbolic tangent (*tanh*) tuning curve to perform a nonlinear operation of its input and can be easily implemented with a few transistors. Our digital FPGA system uses a 'broken-stick' nonlinearity (Fig. 8) [11], which is similar to rectified linear unit (ReLU) activation function often used in deep neural networks [14]. Using the SOUL algorithm, the output weights that describe a linear relationship between the hidden layer and the output layer can be learnt to approximate a desired function as a regression solution, or to perform a classification task for the input-output relationship in the ELM framework.

## III. Opium Learning Algorithm

In a previous work, we have presented an incremental method called Online Pseudo Inverse Update Method (OPIUM) to calculate the pseudoinverse solution of the weight optimisation problem in an online fashion [8]. The solution to the network weights is calculated in a manner similar to the batch method of singular value decomposition (SVD) method [15] often used to calculate the pseudoinverse.

*Algorithm:* Let us consider a three layer feed-forward neural network, having $L$ number of hidden neurons, as shown in Fig. 1. Let $g(.,.,.)$ be a real-valued function so that $g(w_i^{(1)}x + b_i^{(1)} + o_i^{(1)})$ is the output of the $i^{th}$ hidden neuron with random bias $b_i^{(1)} \in \mathbb{R}$ corresponding to the input vector $x \in \mathbb{R}^m$ and the randomly determined weight vector $w_i^{(1)} = (w_{i1}^{(1)},\ldots w_{im}^{(1)})$, where $w_{is}^{(1)}$ is the weight of the connection between the $i^{th}$ hidden neuron and $s^{th}$ neuron of the input layer. Systematic offset $o_i^{(1)} \in \mathbb{R}$ is added to ensure that each neuron exhibits a distinct tuning curve. The output function of the network $f(.)$ is given by:

$$f(x) = \Sigma_{i=1}^{L} w_i^{(2)} g(w_i^{(1)} x + b_i^{(1)} + o_i^{(1)}) \quad (1)$$

where, $w_i^{(2)} = (w_{1i}^{(2)},\ldots w_{ki}^{(2)}) \in \mathbb{R}^k$ is the weight vector where $w_{ji}^{(2)} \in \mathbb{R}$ is the weight connecting the $i^{th}$ hidden neuron with the $j^{th}$ neuron of the output layer. For the $n^{th}$ training pattern, the $i^{th}$ hidden neuron activation is given by:

$$h_{in} = g(w_i^{(1)} x_n + b_i^{(1)} + o_i^{(1)}) \quad (2)$$

where $h_n \in \mathbb{R}^{L \times 1}$. The output weights between $L$ hidden neurons and $K$ output layer nodes [8], $w_n^{(2)} \in \mathbb{R}^{K \times L}$ can be updated as:

$$w_n^{(2)} = w_{n-1}^{(2)} + (y_n - w_{n-1}^{(2)} h_n)\phi_n \quad (3)$$

where, $\phi_n \in \mathbb{R}^{1 \times L}$ is given by:

$$\phi_n = (h_n^T \theta_{n-1})/(1 + h_n^T \theta_{n-1} h_n) \quad (4)$$

$\theta_n \in \mathbb{R}^{L \times L}$ is the inverse of the autocorrelation matrix of the hidden layer activation.

$$\theta_n = \theta_{n-1} - \theta_{n-1} h_n \phi_n \quad (5)$$

This algorithm is equivalent to the Recursive Least Squares algorithm (without forgetting factor) used in adaptive filters [16]. In the following section, we derive the SOUL algorithm from the OPIUM algorithm, by introducing some simplifications to make it hardware friendly.

IV. SIGN-BASED ONLINE UPDATE LEARNING (SOUL) ALGORITHM

In the ELM framework, the input is randomly projected to higher dimensions using random weights, random biases and controllable systematic offsets, and then nonlinearly transformed using the hidden neuron activation function. In a digital system, random input weight vectors can be generated using the LFSR logic [17], while in an analogue system they may arise due to random mismatch of transistors [9]. All these operations ensure that the tuning curves of the hidden neurons across all input patterns are not correlated to each other and that the output range of all the hidden neurons is very similar. This led us to propose a simplification of the OPIUM algorithm, in which we simply assume that the autocorrelation matrix, $\theta_n^{-1} = \varepsilon I$, where $I$ is the identity matrix and $\varepsilon$ is a small positive value. This makes the algorithm provide an inexact, but useful solution to the pseudoinverse and means that we never need to calculate $\theta_n$. The algorithm then simplifies to:

$$\phi_n = h_n^T/(\varepsilon + h_n^T h_n) \quad (6)$$

and, $$w_n^{(2)} = w_{n-1}^{(2)} + (y_n - w_{n-1}^{(2)} h_n)\phi_n \quad (7)$$

$$= w_{n-1}^{(2)} + e_n \phi_n \quad (8)$$

where, $e_n$ is the estimation error. This algorithm is equivalent to the ε-Normalised Least Mean Squares algorithm in adaptive filtering [18], where $\varepsilon$ plays the role of the regularisation constant.

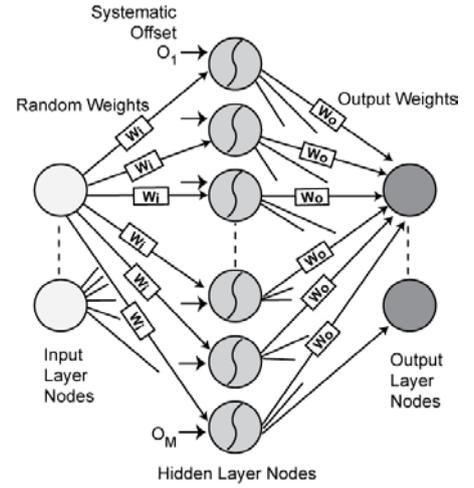

Fig. 1. Architecture of the ELM framework with systematic offset. Three layers of neurons are connected in a feed-forward structure. The connections from the input layer neurons/nodes to the non-linear hidden neurons are via random weights and controllable systematic offsets, $O_1$ to $O_M$. The hidden layer neurons are connected linearly to the output layer neurons via trainable weights. The output neurons compute a linearly weighted sum of the hidden layer values. Adapted from [9].

In (6), the denominator is a scalar quantity. For each new input pattern, the denominator changes due to $h_n$ and it acts as a variable gain and normalising factor, so we can write (6) as:

$$\phi_n = h_n^T/N_{var} \quad (9)$$

The normalising factor, $N_{var}$ in (9), is a variable and would have to be calculated for each connection. The normalising factor influences the convergence time of the output error. We have investigated this option, as it provides, in theory, the best result in terms of the desired input-output mapping. However, it is likely to lead to a circuit per output weight that is too large to be practical, particularly as an accurate analogue-to-digital converter (ADC) is needed for each weight update. A further simplification for the hardware implementations uses a constant gain:

$$\phi_n = h_n^T/N \quad (10)$$

so that (8) reduces to:

$$w_n^{(2)} = w_{n-1}^{(2)} + e_n(h_n^T/N) \quad (11)$$

$w_n^{(2)} \in \mathbb{R}^{K \times L}$ corresponds to the output weights between $L$ hidden neurons and $K$ output layer nodes. In the case of a single output layer node, $w_n^{(2)} \in \mathbb{R}^L$ and $e_n \in \mathbb{R}$ would be scalar.

This stochastic-gradient version is equivalent to the basic Least Mean Squares (LMS) method used in adaptive filtering [18], with $1/N$ representing the learning rate parameter.

Even with these simplifications, (11) requires multiplications of two variables ($e_n$, $h_n$) for each weight connection, which would still need a large circuit per connection. Instead, we simplify the weight update rule even further, by simply incrementing or decrementing the weight by a small amount depending on the product of the sign of the error $e_n$, and the sign of the hidden layer activation $h_n$. This yields the SOUL rule:

$$w_n^{(2)} = w_{n-1}^{(2)} + \text{sign}(e_n)\,\text{sign}(h_n^T)/N \quad (12)$$

and is equivalent to sign-sign LMS in adaptive filtering [18]. This rule only needs a single bit multiplication and can be implemented with an XOR gate.

In the following section, we discuss the hardware implementation of the SOUL algorithm.

## V. HARDWARE IMPLEMENTATION OF THE LEARNING ALGORITHM

The algorithm is implemented via a Digital Learning Block (DLB) developed using digital standard cells for both analogue and digital systems. The DLB calculates and updates the magnitude and polarity of the weight for each new training pattern. There is a DLB block for each connection between a hidden neuron and an output neuron.

A negative sign is represented by '1', and a positive sign by '0'. Thus the product of the signs in (12) can be easily implemented with a standard logical *XOR* gate. The digital weights can then be stored in a counter that counts up or down one unit depending on the product of the signs. This implicitly defines the normalisation factor $N$ as $2^n$, where $n$ is the number of bits used for the counter. This counter effectively stores the magnitude of the weight (*magW*) and the sign of the weight (*signW*). To control the magnitude of the increment/decrement, a 3-bit register (*add_no*) is used, which can take any value from 0 to 7. It allows us to achieve a trade-off between resolution and the learning rate. A high value of *add_no* will result in fast learning but with low resolution weights, while a small value of *add_no* will give a higher resolution in representing the output function but will increase the learning time. A useful option is to decrease the value of *add_no* during training, which will reduce the training time while maintaining a better final resolution.

The pseudocode for the SOUL algorithm is below.

*BEGIN*
*add_count = 1<<add_no;*
   *(@ each posedge clk)*
   *decr = signE $\oplus$ signH;*
  *If (signW $\oplus$ decr):*
       *magW = magW - add_count;*
  *Else:*
       *magW = magW + add_count;*
*END*

where, $\oplus$ represents the *XOR* operation and << represents the left shift operation. Note that the weight changes at each training cycle.

## VI. REGRESSION RESULTS

We have tested the analogue TAB system for regression task with 13 bit weights. The TAB is based on the ELM framework and is described in detail along with circuit descriptions in our previous work [9]. The hidden neurons of the TAB use a hyperbolic tangent (*tanh*) tuning curve to perform a nonlinear operation of its input and can be easily implemented with a few transistors (Fig. 7A). The hidden and output layers are connected via DLB controlled output weight circuit (Fig. 7B). Here, we present the simulated learning results of the TAB framework using Python and circuit simulations.

### A. Numerical (Python) simulations

We configured the TAB framework for a single input and a single output (SISO) and a hidden layer of 100 neurons. We modelled the input of the TAB as a voltage variable and the output as a current variable in the range of nanoamperes. The TAB architecture was trained for various functions ($Y = f(X)$), ranging from simple functions such as *cube* and *sine*, to complex functions such as *sinc*. For the simulations, the input range was normalised between [-1, 1] V, and was quantised using 200 points. During training, we presented the (*X, Y*) training pairs in random order and repeated this for a defined number of training epochs. For example, 10 epochs would mean presenting 10x200 = 2000 iterations. Upon the completion of training, we presented only the input to the system and observed the learned output. We verified the learning performance by calculating the error as the difference between the observed output and the predicted output. Fig. 2 shows the learned output of the TAB in the testing phase for various regression tasks, as well as the target output and the error. Additionally, we show the convergence of the error in the training phase, which depends on the complexity of the function. Empirically, the more changes in the sign of the derivative of the function (more wiggles), the more time it takes to minimise the error. Convergence of the error (Fig. 2, Ib-IIIb) was analysed by keeping track of the number of training examples needed to reach a pre-determined level of accuracy.

*1) Error vs number of bits in the output weight:* The RMS error as a function of the number of bits used for the output weights, for $y = \text{sinc}(6\pi x)$ for a network with 100 hidden nodes is shown in Fig. 3. Again, these are the results of 10 simulations with different random weights from the input to the hidden nodes. From about 8-bits onwards, the variance is negligibly small, and the RMS (Root Mean Square) error becomes almost totally independent of the random weights. Increasing the number of bits per weight is a matter of diminishing returns, and 11-bits seem sufficient, even to learn this difficult function.

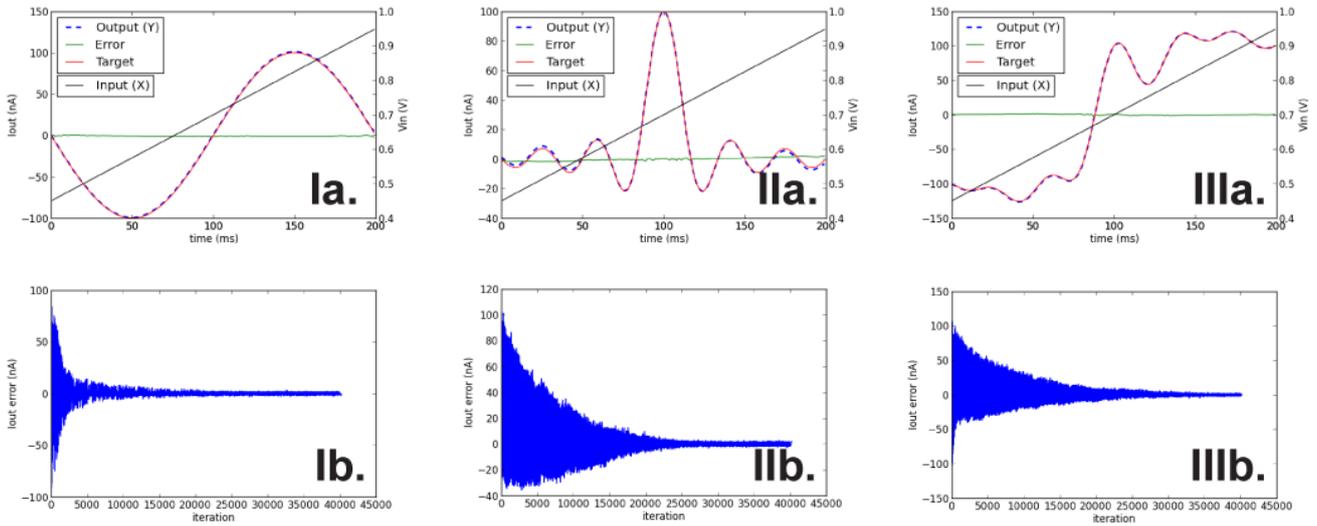

**Fig. 2. Python simulations of the TAB for a SISO system.** The TAB system with a hidden layer of 100 neurons was trained to learn **(I)** *Sine*, **(II)** *Sinc*, and **(III)** a *Complex* function ($Y = sin(x) + x^3 + sin(x)/x$) using online supervised learning with a training dataset containing a set of samples, where each sample was a pair consisting of an input ($V_{in}$, *black line*) and a desired output value ($I_{out}$, *red curve*). In **(a)**, the learned function (*dashed curve*) approximates the correct output value (*red curve*) for the test input and the error (*green curve*) is close to 0. In **(b)**, the error converges to a minimum over the learning process.

*2) Stochasticity of training results in better performance:* Stochasticity in the SOUL algorithm arises from the random presentation of training data to the TAB. We compared the performance of the learning algorithm for drawing training samples in an ordered manner versus in a random manner for a fixed number of training iterations. We trained the TAB to learn the *sinc* function (with RMS = 28.6) for different numbers of hidden neurons, and compared the percentage RMS error with the target RMS for both the cases. As shown in Fig. 4, random shuffling of the training pairs results in a better learning performance as compared to no shuffling. The difference in percentage error becomes significantly pronounced when the number of hidden neurons is high (>60).

*3) Error convergence time depends upon the number of hidden neurons and training epochs:* We have observed in our simulations that the number of epochs required to converge to a given error threshold is inversely proportional to the number of hidden neurons in the circuit. In addition, the TAB requires a minimum number of hidden neurons to encode the input stimulus in order to be able to converge to a given error threshold. Similarly, there is a minimum number of epochs required for error convergence, and this is not affected by increasing the number of hidden neurons. Fig. 5 shows the number of iterations versus the number of hidden neurons for the *sinc* function. We have used an error threshold of 3% (RMS for *sinc* function = 28.6), and measured the total number of epochs required to reach this threshold with the given number of hidden neurons. In Fig. 5, the red dot shows the minimum number of hidden neurons required as 26, below which we cannot achieve accuracy within the given threshold.

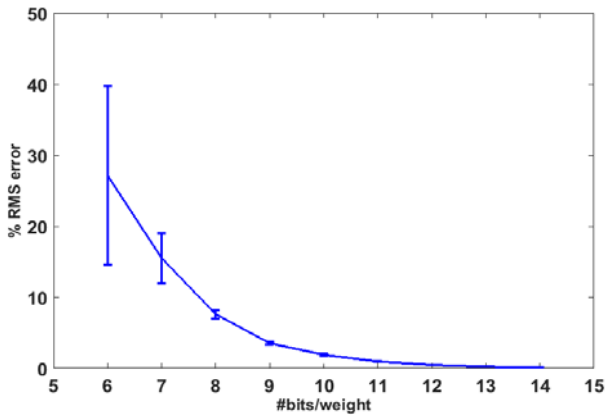

**Fig. 3.** The percentage RMS error versus the number of bits/weight used for the output weights for the function $y = sinc(6\pi x)$. The error bars show the standard deviation.

*4) Variable step size:* The SOUL algorithm provides an additional option to change the step size in the learning rule, which helps to converge to the minimum error at a much faster rate. In Fig. 6, we have compared the learning performance with a constant step size versus a variable step size for the *sinc* function. We can change the value of *add_no* during the training (pseudocode, Section V), which will reduce the training time significantly. In Fig. 6, the *add_no* for the variable step size was changed from 3 (up to iteration 2000) to 0 (from iteration 2000 till last). This resulted in the error converging much faster than using a fixed step size (w.r.t. *add_no* = 0 i.e add_count = 1, as in pseudococde), while achieving the same accuracy.

Overall, the simulation results of the training suggest that the system can be successfully trained to perform various regression tasks. Also, we have identified key features for optimal performance of the SOUL algorithm in a TAB.

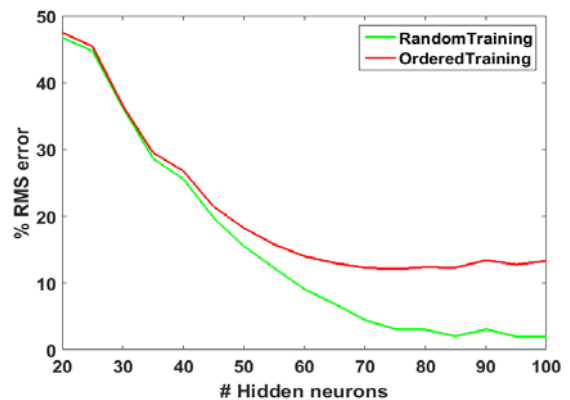

**Fig. 4. Plot of percentage error as a function of the number of hidden neurons, for random and ordered training.** Training data were presented to the TAB with a varying number of hidden neurons in a random or ordered manner, and the errors were compared.

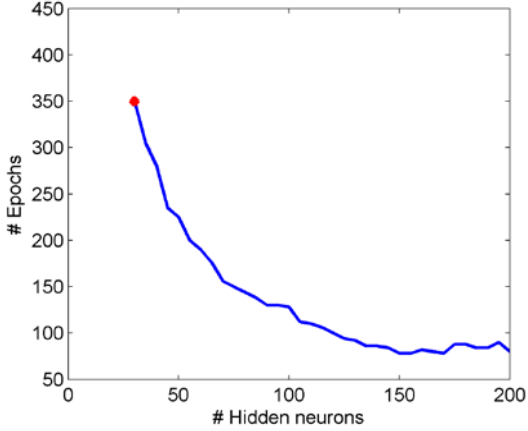

Fig. 5. Plot of the number of epochs versus the number of hidden neurons for the *sinc* function. Red dot represents the minimum number of hidden neurons required to achieve an error threshold of 3%.

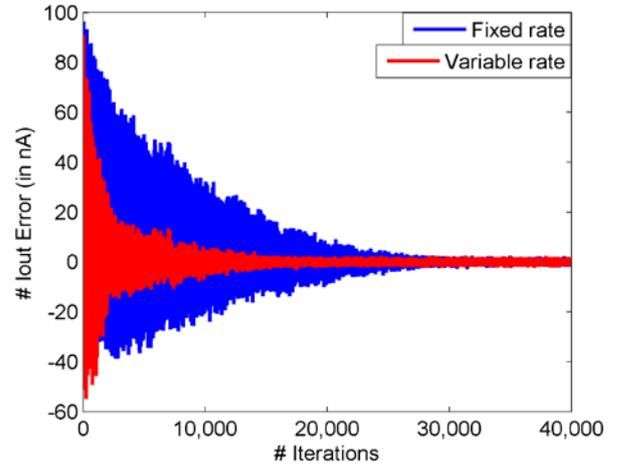

Fig. 6. Error evolution for the *sinc* function using a fixed update rate and a variable update rate using the SOL algorithm in a TAB. The *add_no* variable (as explained in section V) for the variable rate was changed from 3 to 0 at iteration 2000.

*B. Circuit Simulations*

After obtaining encouraging results from the Python simulations, we implemented the TAB framework (Fig. 1) using transistor circuits. We implemented the hidden neuron *tanh* model using a differential pair circuit, and the output weight logic using a splitter circuit (Fig. 7A, B). The splitter circuit is used along with the DLB to implement the SOUL algorithm in the TAB system. A detailed description of the hidden neuron and splitter circuits is given in our previous work [9]. The DLB is explained in detail in section V. For each training pattern, the DLB corresponding to each hidden neuron generates a 13-bit binary number. Each bit of this binary number is used as a binary switch, which controls the current in each splitter branch. We have used 65nm transistor models for transistor level circuit simulations. In the simulation, there are a total of only 20 hidden neurons and each neuron exhibits a different systematic offset $V_{ref}$, which results in a different nonlinear tuning curve for each neuron and is thus useful for learning. The SOUL method uses the sign of the hidden neuron, which is shown in Fig. 7A as the *signH* port. For each training pattern presented to the system at a time, a predicted output is computed. Based on the target and the predicted output, the sign of the error signal feeds back to the DLB corresponding to each hidden neuron. Based on the sign of the error (*signE*) and the sign of each hidden neuron (*signH*), the output weight corresponding to each hidden neuron gets updated for each training pattern. We have trained the TAB network in a single input and a single output configuration with 20 hidden neurons to learn the *cube* function. The TAB is trained only for a few iterations due to the extremely long simulation time (it takes 10 hours to run a simulation of 10 ms) and the fact that the memory requirement grows linearly with simulation time. During the final 0.4 ms, training was turned off and only the input was presented to the circuit to test how well the function was learned. Much of the 'noise' in the first 10 ms (Fig. 7C) is thus the result of changes in the output weights during training. This circuit simulation shows that the TAB can be successfully implemented using transistor circuits and trained for various learning tasks, but it is likely that more than 20 hidden neurons will be needed to learn this function, and more than 25 epochs to fully train the circuit.

## VII. CLASSIFICATION RESULTS

We have previously developed a digital system for offline learning using this framework [11]. We employed the same system here to apply the SOUL algorithm and we have verified the performance of the SOUL algorithm for digit recognition tasks on the MNIST database [12]. In the MNIST database, the digits are represented as 28×28 = 784 pixels, and the training and the testing datasets contain 60,000 and 10,000 digits, respectively. We have employed our digital system, called **Neu**romorphic **P**attern-recognition **S**ystem (NeuPS), which is based on the ELM principles and is similar to the TAB framework, but implemented on FPGA. Currently, NeuPS has been developed for binary images, so we first converted the MNIST greyscale images into binary images. The NeuPS is a three-layer feed forward neural network, consisting of 784 input layer neurons (pixels), and a large number of hidden layer neurons and ten output layer neurons. In the description below, we will use 8192 hidden neurons for simplicity, however, we have characterised the results for various numbers of hidden neurons in the result section. The input layer neurons are connected to the hidden layer neurons using randomly weighted all-to-all connections. The random weights in the NeuPS are generated on-the-fly using an LFSR pseudo random number generation circuit. The hidden layer neurons are also connected to the output layer neurons using all-to-all connections, but with output weights calculated using the SOUL learning algorithm. In the digit recognition system, a single $n^{th}$ input digit (28×28=784 pixels) is mapped onto a layer of input neurons, which we refer to as vector $x_n \in \mathbb{R}^{784 \times 1}$. The random input weight matrix is $w_{rnd}^{(1)} \in \mathbb{R}^{8192 \times 784}$. The input to the hidden neurons for an $n^{th}$ input digit, $v_n \in \mathbb{R}^{8192 \times 1}$, is thus given by:

$$v_n = w_{rnd}^{(1)} x_n$$

Each value in $v_n$ is the sum of the randomly weighted pixels, and is the stimulus for the corresponding neuron in the hidden layer. The tuning curve of the hidden layer neurons are implemented using 'broken-stick' nonlinearity and discussed in the next section (Fig. 8). The decoding weights $w^{(2)} \in \mathbb{R}^{10 \times 8192}$ are updated according to the SOUL rule and the output vector $y \in \mathbb{R}^{10 \times 1}$ represents the predicted value of the input digit.

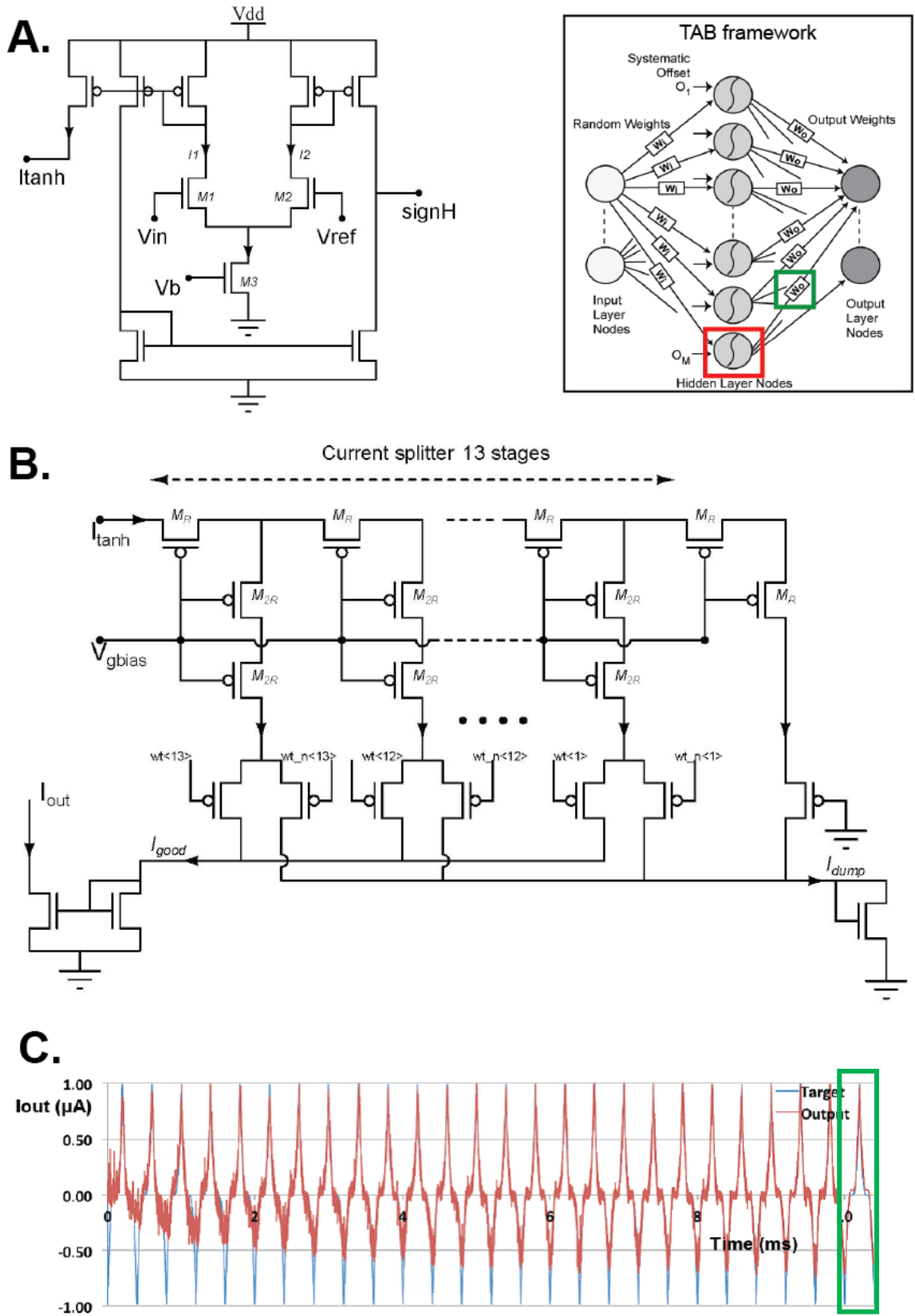

**Fig. 7. Circuit simulation of the TAB circuit to learn $y = x^3$. A.** Hidden neuron implements *tanh* nonlinearity. Each hidden layer node (*red box*) in the TAB framework is represented by this circuit. **B.** Splitter circuit, which is used along with a digital logic block to implement the SOL algorithm. Each weight block (*green box*) of the TAB framework consists of the splitter circuit and a digital logic block (not shown). **C.** Simulation results. x-axis shows the learning time in ms (millisecond) and y-axis shows the output current in μA (microampere). At 10 ms, training was turned off and then learning was tested in the last 0.4 ms (*green box*). Noise in the first 10 ms is a result of changes in the output weights during training.

The above description is for a single digit. Here, we used 60,000 sample digits in the training phase and 10,000 sample digits in the testing phase. In the testing phase, the predicted output $Y$ is compared with the expected output to obtain the error rate (i.e. the number of digits recognised wrongly among 10,000 test digits). To efficiently implement the NeuPS on an FPGA, we use a time-multiplexing approach [19]–[23], which leverages the high-speed digital circuit. State-of-the-art FPGAs can easily run at a clock speed of 266MHz (clock period 3.75ns). Here, we have modelled the NeuPS FPGA into MATLAB using fixed points, and all the analysis and results reported here are from the fixed point MATLAB model of the NeuPS.

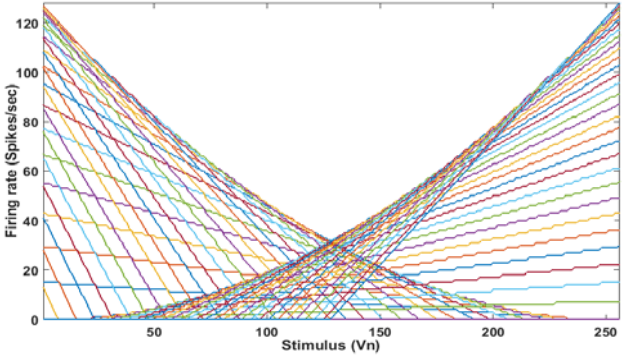

**Fig. 8** The tuning curves of 64 hidden neurons in the NeuPS.

*A. Hidden neuron tuning curves*

The hidden neuron in the TAB framework implements the *tanh* nonlinearity, which is hardware intensive to implement using digital circuits. Instead, in the NeuPS, 'broken-stick' nonlinearity has been implemented as a hidden layer activation function. Fig. 8 shows the tuning curves of a set of 64 hidden layer neurons. The transfer function is thus a nonlinear function of its input, since the value of the output of neurons cannot be negative. Instead, this piece-wise linear function can be easily implemented using a single 9-bit fixed-point multiplier. Since the time-multiplexing approach has been used, one physical neuron can simulate 256k neurons in one millisecond in the NeuPS [11].

*B. Results*

We have investigated the accuracy of the NeuPS for different values of bit size of the output weights and number of hidden neurons. As the number of hidden layer neurons grows, the total number of output weights will grow and subsequently the memory storage of these weights will increase. Thus, we trained the NeuPS for bit size ranging from 13 bits to 17 bits, but always used only the 6 Most Significant Bits (MSB) once the system was trained. This leads to a significant savings in memory use.

The SOUL algorithm uses a fixed step size (normalising factor *N*), which influences the convergence time. For a large *N*, the convergence time would be longer, thus it might need more training patterns to reach the maximum accuracy for a given number of hidden neurons. In contrast, a smaller *N* will decrease the convergence time, but the system may not achieve the best accuracy for that number of hidden neurons. In Fig. 9, we show the results obtained from testing the NeuPS by varying the output weight bits resolution (normalising factor *N*), the number of hidden neurons and the number of training epochs (one training epochs = 60000 patterns). The best result of 95.05% is achieved with 16384 ($2^{14}$=16*K*) hidden neurons with 15 bits output weight bits during training and 3 training epochs of the database. In Fig. 10, we show the mean of 50 different seeds of random weights for the curve corresponding to the best value of normalising factor of Fig. 9 for different numbers of hidden neurons. The SOUL algorithm achieves the best trade-off between accuracy and feasibility of hardware implementation. In a real world application where the amount of training data is large, an online learning algorithm such as the SOUL algorithm, which is simple and area efficient, is a good choice for 'Big Data' applications.

## VIII. CONCLUSIONS

We have derived and developed a sign-based online update learning (SOUL) algorithm based on the OPIUM algorithm. We have tested the performance of the SOUL for MNIST digit recognition tasks. Also, we have analysed and trained an analogue TAB for various regression tasks and observed the evolution of error. Our analysis demonstrates satisfactory learning performance using the SOUL algorithm with a very simple hardware implementation. We have also analysed the relation between the number of hidden nodes and the convergence of error. We have also implemented a TAB in a single input and single output configuration using transistors, and shown that it learns successfully using the SOUL method. Our proposed SOUL algorithm minimises a cost function based on a least squares solution and may find applications in adaptive hardware neural architectures. The speed and power consumption of today's CPUs/GPUs are greatly challenged by the need of information analytics on big data platforms. Recent advances in sensing technology further aggravate the situation with a big volume of data. Therefore, power and area efficient algorithms similar to the SOUL on custom hardware are essential to accelerate learning and classification on a chip under stern area and power constraints.

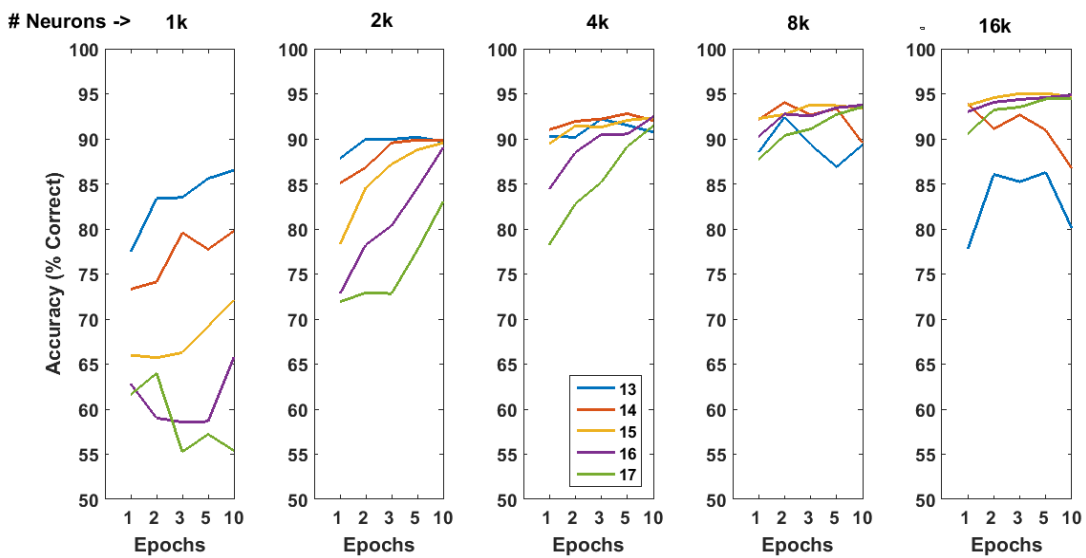

**Fig. 9.** Accuracy as a function of the number of hidden layer neurons, training epochs and output weight bit size.

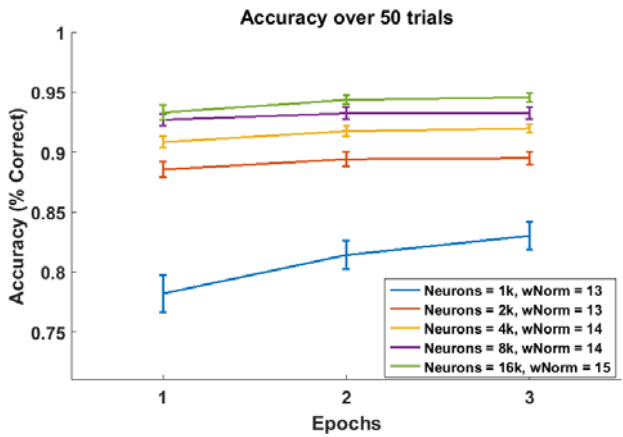

**Fig. 10.** Mean accuracy over 50 trials as a function of hidden neurons and wNorm (normalising factor).